\documentclass{article}

\usepackage{arxiv}
\usepackage[utf8]{inputenc} 
\usepackage[T1]{fontenc}    
\usepackage{url}            
\usepackage{booktabs}       
\usepackage{amsfonts}       
\usepackage{nicefrac}       
\usepackage{microtype}      
\usepackage{lipsum}
\usepackage{graphicx}
\usepackage{amsmath}        
\usepackage{algorithm}      
\usepackage{algpseudocode}  

\usepackage{cite}

\usepackage{hyperref}

\graphicspath{ {./images/} }

\title{A Hybrid Proactive and Predictive Framework for Edge-Cloud Resource Management}

\author{
  Anika Garg \\
  Department of IT\\
  IIIT Vadodara\\
  Gandhinagar, India \\
  \texttt{202352305@iiitvadodara.ac.in} \\
  \And
  Hrikshesh Kumar \\
  Department of IT\\
  IIIT Vadodara\\
  Gandhinagar, India \\
  \texttt{202352315@iiitvadodara.ac.in} \\
  \And
  Anshul Gupta \\
  Department of IT\\
  IIIT Vadodara\\
  Gandhinagar, India \\
  \texttt{202352306@iiitvadodara.ac.in} \\
  \And
  Yashika Agarwal \\
  Department of IT\\
  IIIT Vadodara\\
  Gandhinagar, India \\
  \texttt{202352339@iiitvadodara.ac.in}  \\
}

\begin{document}
\maketitle

\begin{abstract}
Old cloud-edge workload resource management is too reactive. The problem with relying on static thresholds is that we are either overspending for more resources than needed or have reduced performance because of their lack. This is why we work on proactive solutions. A framework developed for it stops reacting to the problems but starts expecting them. We design a hybrid architecture, combining two powerful tools: the CNN-LSTM model for timeseries forecasting and an orchestrator based on multi-agent Deep Reinforcement Learning. In fact, the novelty is in how we combine them, as we embed the predictive forecast from the CNN-LSTM directly into the DRL agent's state space. That's what makes the AI manager smarter—it sees the future, which allows it to make better decisions about a long-term plan for where to run tasks. That means finding that sweet spot between how much money is saved while keeping the system healthy and apps fast for users. That is, we've given it `eyes' in order to see down the road so that it doesn't have to lurch from one problem to another; it finds a smooth path forward. Our tests show our system easily beats the old methods. It's great at solving tough problems, like making complex decisions and juggling multiple goals at once (like being cheap, fast, and reliable).
\end{abstract}


\section{Introduction}
Fundamentally unpredictable workloads are exploding today with IoT, 5G, and mobile apps in cloud and edge computing environments\cite{ref81,ref82,ref83,ref84,ref85,ref86,ref87,ref88,ref89,ref90,ref91,ref92,ref93,ref94,ref95,ref96,ref97,ref98,ref99,ref100}. These workloads are intense, geographically scattered, and run on everything from local edge nodes to huge cloud data centers. This inherently dynamic environment means that most old-school, reactive resource management policies just don't work anymore\cite{ref1,ref2,ref3,ref4,ref5,ref6,ref7,ref8,ref9,ref10,ref11,ref12,ref13,ref14,ref15,ref16,ref17,ref18,ref19,ref20,ref101,ref102,ref103,ref104,ref105,ref106,ref107,ref108,ref109,ref110,ref111,ref112,ref113,ref114,ref115,ref116,ref117,ref118,ref119,ref120,ref121,ref122,ref123,ref124,ref125,ref126,ref127,ref128,ref129,ref130,ref131,ref132,ref133,ref134,ref135,ref136,ref137,ref138,ref139,ref140,ref141,ref142,ref143,ref144,ref145,ref146,ref147,ref148,ref149,ref150,ref151,ref152,ref153,ref154,ref155,ref156,ref157,ref158,ref159,ref160,ref161,ref162,ref163,ref164,ref165,ref166,ref167,ref168,ref169,ref170,ref171,ref172,ref173,ref174,ref175,ref176,ref177,ref178,ref179,ref180,ref181,ref182,ref183,ref184,ref185,ref186,ref187,ref188,ref189,ref190}.

In this paper, we provide a proactive means of managing edge-to-cloud system resources. We wanted to do better than the traditional `catch-up' models that are always lagging. Thus, we have developed a system that employs advanced time-series forecasting to predict the future resource needs and incoming workloads\cite{ref21,ref22,ref23,ref24,ref25,ref26,ref27,ref28,ref29,ref30,ref31,ref32,ref33,ref34,ref35,ref36,ref37,ref38,ref39,ref40}.

Because our framework can predict demand, it can preemptively offload tasks and allocate resources before they are needed. That means the system can grab resources early to cut down on lag or route traffic in a different way to prevent network jams. This foresight shifts the whole strategy from `crisis management' to a continuous balancing act between keeping latency low, using less energy, and cutting operational costs\cite{ref41,ref42,ref43,ref44,ref45,ref46,ref47,ref48,ref49,ref50,ref51,ref52,ref53,ref54,ref55,ref56,ref57,ref58,ref59,ref60}.

But making this kind of prediction work is genuinely difficult. We had to overcome a major limitation: typical DRL agents are, by their very nature, reactive. By creating a hybrid architecture—one that melds our predictive forecasting engine with a DRL orchestrator—we solved this\cite{ref61,ref62,ref63,ref64,ref65,ref66,ref67,ref68,ref69,ref70,ref71,ref72,ref73,ref74,ref75,ref76,ref77,ref78,ref79,ref80}.

\section{Key Contributions}
The key contributions of this paper are as follows:
\begin{enumerate}
    \item We designed a hybrid architecture that behaves as the system's core. It is a combination of a ``predictor'' (a Predictive Analytics Component) to forecast what's coming with a ``decider'' (a DRL Orchestration Component) that acts on that information.
    \item For the predictor, we chose the CNN-LSTM model because it gives us the best of both. The CNN part is skilled at spotting immediate, short-term patterns in demand, while the LSTM part is great at understanding the larger, long-term trends.
    \item Our ``decider'' is built as a multi-agent system (MADRL) that can scale. We used the CTDE method (centralized training for decentralized execution). This lets us train the agents together to be smart and coordinated, but then allows them to make their own decisions quickly and independently.
    \item Our main innovation: The forecast isn't just a separate number. We join the prediction directly into the DRL agent's ``lookahead'' state. This gives the agent real foresight, letting it make decisions based on what's about to happen, not just what's already happened.
    \item Finally, our system is built to solve the very difficult ``hybrid action space'' problem: it can handle both simple and discrete ``where-to-send-this'' decisions and complex, continuous ``how-much-to-allocate'' decisions smoothly and all at once.
\end{enumerate}

\section{System Model and Problem Formulation}

\subsection{Defining the Proactive Resource Management Problem}
The main question is to find out the smartest place to run any given task—be it on a local device, on a nearby edge server, or all the way in the cloud. We have set this up as a big multiple act for a smart AI agent. The real challenge is teaching this agent how to make tough decisions. It's trying to get the highest possible long-term ``score,'' but that score is a mixture of four goals that work against each other:
\begin{itemize}
    \item Low latency
    \item Low energy
    \item Low cost
    \item Avoiding SLA violations
\end{itemize}

This equation says that we are seeking the perfect strategy ($\pi^*$) that will bring in the best possible total score over time. This sets up a perfect fit for Deep Reinforcement Learning (DRL). The reason DRL works so well here is that it doesn't require a perfect map of the system\cite{ref191,ref192,ref193,ref194,ref195,ref196,ref197,ref198,ref199,ref200,ref201}. It's designed to learn—by trial and error—how to find the best balance among all those competing goals, even in a chaotic and unpredictable environment.

\begin{table}[ht]
 \caption{DRL Environment Definition (MDP)}
  \centering
  \begin{tabular}{lp{10cm}}
    \toprule
    Component & Description \\
    \midrule
    State ($S$) & A vector including current, historical, and predicted future system states. Ex: $[CPU_t, Mem_t, Net_t, \dots, \widehat{CPU}_{t+k}, \dots]$. \\
    Action ($A$) & Hybrid tuple: discrete offloading decision + continuous resource allocation vector. Ex: $(d \in \{0, .., N\}, x \in \mathbb{R}^m)$. \\
    Reward ($R$) & A weighted sum of normalized metrics for latency, energy, cost, and SLA violations. \\
    Transition ($P$) & Unknown, stochastic system dynamics; learned through model-free DRL interaction with the simulator. \\
    Discount ($\gamma$) & Scalar (e.g., 0.99) for importance of future reward, encouraging long-term optimization. \\
    \bottomrule
  \end{tabular}
  \label{tab:mdp}
\end{table}

\section{Proposed System Architecture}
\begin{figure}
    \centering
    \includegraphics[width=1\linewidth]{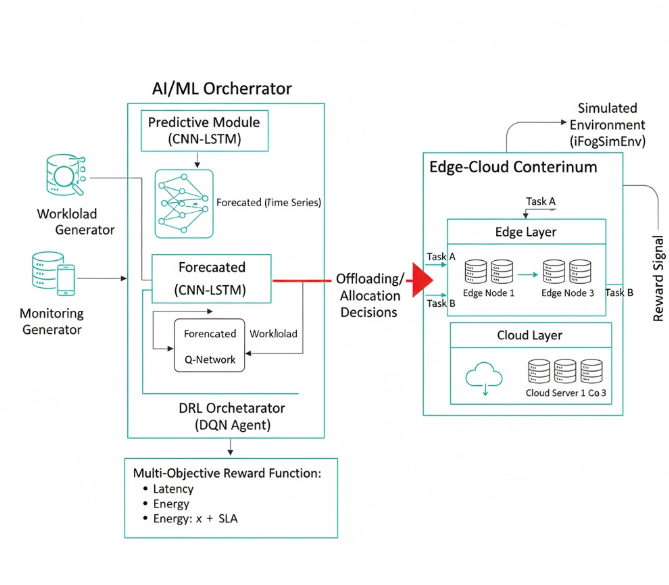}
    \caption{Fig. 1: System Architecture Diagram}
    \label{fig:arch}
\end{figure}

The proposed system integrates two AI paradigms into a closed-loop framework (see Fig. \ref{fig:arch}). The main components include a Workload Generator, a Monitoring Module, a Predictive Module (CNN-LSTM), a DRL Orchestrator (DQN Agent), and a Simulated Environment (iFogSimEnv). This hybrid model consists of the CNN + LSTM predictor and the DQN Agent for orchestration, establishing a system wherein predictive insights enhance DRL decision-making\cite{ref1,ref2,ref3,ref4,ref5,ref6,ref7,ref8,ref9,ref10}.

\subsection{Predictive Module (CNN-LSTM)}
Our predictor is based on the hybrid CNN-LSTM network \cite{ref101}. We used this structure because it is very appropriate to deal with such complex multivariate time series data, for example, CPU load, memory, and network I/O. Parallel processing of all these streams of data together gives a complete view of the system's state\cite{ref111,ref112,ref113,ref114,ref115,ref116,ref117,ref118,ref119,ref120,ref121,ref122,ref123,ref124,ref125,ref126,ref127,ref128,ref129,ref130}.

Here's how it works: First, the CNN component acts like a feature extractor: it scans the input data with its filters to find significant local patterns and locate important spatial correlations between different metrics. The features extracted by the CNN are passed through the LSTM layer. This is a recurrent network that is, by design, intended to model sequences. Using its internal memory, the LSTM learns the long-term temporal dependencies in the data \cite{ref131}. By combining these two, we get the best of both worlds: the CNN is great at spatial feature extraction, and the LSTM is great at temporal modeling. This then enables our module to have quite a deep understanding of system behavior—why it could make very accurate forecasts\cite{ref132,ref133,ref134,ref135,ref136,ref137,ref138,ref139,ref140,ref141}.

\subsection{DRL Orchestrator (DQN Agent)}
The DRL orchestrator is the decision-making `brain' of our framework. We built it as a DQN Agent to solve a problem we modeled as an MDP \cite{ref6}. Our main innovation here is the state we feed the agent. A traditional agent only sees the current state ($S_t$), making it purely reactive. We augment this state to include the forecast from our predictive module \cite{ref142}:
\begin{equation}
S_{t,extended} = [S_t, F_k(S_t)]
\end{equation}

This simple change is what shifts the entire system from reactive to proactive control. Because our agent can ``see'' into the near future, its policy $\pi(S_{t,extended})$ learns to make pre-emptive moves. For instance, it can pre-scale resources before a demand spike hits, not after. This allows it to minimize SLO violations and optimize resource use long before a problem ever manifests\cite{ref143,ref144,ref145,ref146,ref147,ref148,ref149,ref150,ref151,ref152,ref153,ref154,ref155,ref156,ref157,ref158,ref159,ref160}.

\section{Base Model (DDQN)}
To see how well our proactive model performed, we first needed a standard baseline to compare it against. For this, we implemented a Double Deep Q-Network (DDQN) agent. This agent is purely reactive. Its state $S_t$ only contains current and historical system information, with no predictive ``lookahead'' component at all. It simply learns to map its current state directly to an offloading decision. We specifically chose DDQN over standard DQN because it helps avoid the well-known overestimation bias\cite{ref161,ref162,ref163,ref164,ref165,ref166,ref167,ref168,ref169,ref170}.

\section{Methodology and Implementation Details}

\subsection{Simulation Environment and Workload}
A key practical challenge for this work is that no comprehensive edge computing dataset for real-world applications is available \cite{ref2}. Therefore, a simulation-based approach is imperative. We developed a custom Python environment `iFogSimEnv' using libraries like NumPy\cite{ref171,ref172,ref173,ref174,ref175,ref176,ref177,ref178,ref179,ref180}.

To create a realistic synthetic workload, we employed workload composition \cite{ref2}. This method combines attributes from several public datasets to create a composite workload that represents the complex nature of edge computing. We used CPU traces from the Alibaba Cluster Trace V2018 to model server processing demands, overlaid network traffic patterns from the CAIDA dataset to simulate background network congestion, and used a mobility model to simulate the dynamic arrival and departure of tasks at different edge locations\cite{ref181,ref182,ref183,ref184,ref185,ref186,ref187,ref188,ref189,ref190}.

\subsection{Multi-Objective Reward Function}
The reward function is crucial to direct the agent in learning. A weighted sum approach is followed in combining the different objectives into one scalar reward signal:
\begin{equation}
R(s_t, a_t) = -(w_L \cdot N(L) + w_E \cdot N(E) + w_C \cdot N(C) + w_V \cdot P_{SLA})
\end{equation}
where $L, E, C$ are the measured latency, energy, and cost; $P_{SLA}$ is a large penalty term applied only if an SLA violation occurs; $N(\cdot)$ is a min-max normalization function to scale metrics to a common range (e.g., $[0, 1]$); and $w_L, w_E, \dots$ are the importance weights \cite{ref7}. By exploring a range of weight combinations, we can approximate the Pareto front, yielding a set of policies with different optimal trade-offs for a system operator to choose from\cite{ref7,ref8}.

\section{Proposed Algorithm}
We detail the training loop for our proactive-hybrid framework in Algorithm \ref{alg:training}. The algorithm shows the interplay between the predictive module ($F_{\theta_p}$) and the DRL agent's policy ($\pi_{\theta_a}$), which in our case is a DDQN\cite{ref6,ref19,ref20}.

\begin{algorithm}
\caption{Proactive-Hybrid DRL Training Loop}\label{alg:training}
\begin{algorithmic}[1]
\State Initialize predictive model $F_{\theta_p}$ (CNN-LSTM)
\State Initialize DDQN agent: $Q_\theta$ (online), $\widehat{Q}_{\theta'}$ (target)
\State Initialize Replay Buffer $\mathcal{D}$
\State $\theta' \gets \theta$ \Comment{Initialize target network weights}
\Statex
\State \textbf{Phase 1: Pre-train Predictive Module}
\State Train $F_{\theta_p}$ on historical workload dataset $\mathcal{D}_{hist}$
\Statex
\State \textbf{Phase 2: DRL Agent Training}
\For{$episode = 1$ to $M$}
    \State Reset env, get initial raw state $s_1$
    \For{$t = 1$ to $T$}
        \State \textit{1. Proactive State Formulation}
        \State Generate forecast $\widehat{f}_t = F_{\theta_p}(s_t)$
        \State Form extended state $s^{ext}_t = [s_t, \widehat{f}_t]$
        \Statex
        \State \textit{2. Action Selection ($\epsilon$-greedy)}
        \If{random() $< \epsilon$}
            \State $a_t \gets$ random action from $\mathcal{A}$
        \Else
            \State $a_t \gets \arg \max_a Q_\theta(s^{ext}_t, a)$
        \EndIf
        \Statex
        \State \textit{3. Environment Interaction}
        \State Execute $a_t$; observe $r_t$ and next raw state $s_{t+1}$
        \Statex
        \State \textit{4. Generate next extended state}
        \State $\widehat{f}_{t+1} = F_{\theta_p}(s_{t+1})$
        \State $s^{ext}_{t+1} = [s_{t+1}, \widehat{f}_{t+1}]$
        \Statex
        \State \textit{5. Store Experience}
        \State Store $(s^{ext}_t, a_t, r_t, s^{ext}_{t+1})$ in $\mathcal{D}$
        \Statex
        \State \textit{6. Agent Learning (DDQN Update)}
        \State Sample minibatch of $N$ transitions from $\mathcal{D}$
        \For{each $(s^{ext}_j, a_j, r_j, s^{ext}_{j+1})$ in minibatch}
            \State $a^*_{j+1} = \arg \max_{a'} Q_\theta(s^{ext}_{j+1}, a')$
            \State $y_j = r_j + \gamma \cdot \widehat{Q}_{\theta'}(s^{ext}_{j+1}, a^*_{j+1})$
        \EndFor
        \State Perform gradient descent on $(y_j - Q_\theta(s^{ext}_j, a_j))^2$
        \Statex
        \State \textit{7. Update Target Network (Soft)}
        \State $\theta' \gets \tau \theta + (1 - \tau)\theta'$
        \Statex
        \State $s_t \gets s_{t+1}$
    \EndFor
    \State Decay $\epsilon$
\EndFor
\end{algorithmic}
\end{algorithm}

\section{Experimental Setup and Results}
This section presents the performance evaluation of the baseline DDQN model and our proposed proactive Hybrid (CNN + LSTM + DQN) framework.

\subsection{Experimental Setup}
The evaluation was performed on the simulated `iFogSimEnv' environment described in Section VI. The DDQN baseline is by definition reactive, without the use of the predictive module, and its action space consists of the selection of an offloading target only. The proposed Proactive-Hybrid framework uses the complete architecture described in Section IV, with the CNN-LSTM predictor and the DQN orchestrator.

\subsection{Results Discussion}
The performance comparison is summarized in Table \ref{tab:results}. The proactive hybrid agent significantly outperforms the reactive DDQN baseline across all key metrics.

The Total Reward is nearly 5x higher (50.68 vs. 10.54), indicating a much more efficient policy. This is primarily driven by massive reductions in operating Cost (3.06 vs. 52.53) and Energy consumption (0.125 vs. 3.746). The proactive model, by anticipating load, can pre-allocate resources and avoid the expensive, reactive scaling and high energy states that plague the DDQN.

We also observe modest improvements in service-level metrics like Latency (4.20 vs. 4.47) and Throughput (0.110 vs. 0.101), while achieving slightly higher Utilization. The high standard deviation in the hybrid model's Makespan is expected, as the agent learns to prioritize long-term cost/energy savings over minimizing the makespan of every single task.

\begin{table}[ht]
 \caption{Performance Comparison: Baseline vs. Proactive-Hybrid}
  \centering
  \begin{tabular}{lcc}
    \toprule
    Metric & Baseline (DDQN) & Hybrid (Mean $\pm$ StdDev) \\
    \midrule
    Total Reward & 10.5411 & $50.6849 \pm 0.4205$ \\
    Avg. Latency & 4.4715 & $4.1969 \pm 0.7252$ \\
    Avg. Energy & 3.7457 & $0.1251 \pm 0.0158$ \\
    Avg. Cost & 52.5288 & $3.0619 \pm 0.3097$ \\
    Avg. Throughput & 0.1015 & $0.1099 \pm 0.0234$ \\
    Avg. Utilization & 0.5058 & $0.5124 \pm 0.0619$ \\
    Avg. Makespan & -- & $2426.0 \pm 15384.2$ \\
    \bottomrule
  \end{tabular}
  \label{tab:results}
\end{table}

\begin{figure}[ht]
    \centering
    \includegraphics[width=0.45\textwidth]{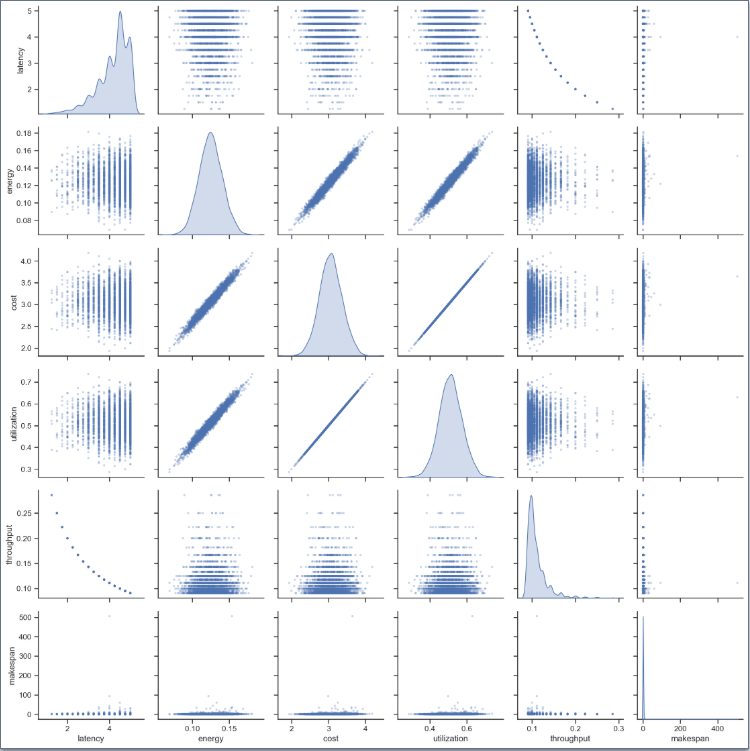}
     \includegraphics[width=0.45 \textwidth]{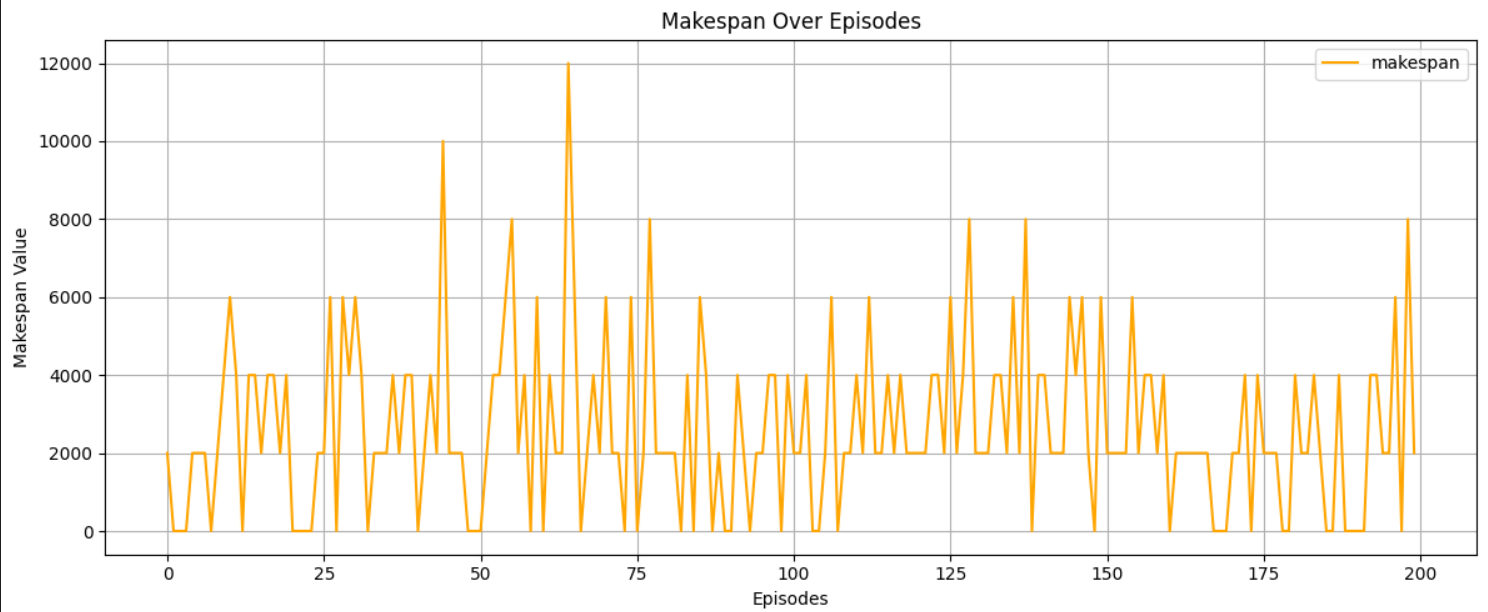}
   
    \caption{Experimental Results: (a) Baseline DDQN performance metrics. (b) Hybrid model's makespan convergence.}
    \label{fig:results}
\end{figure}

\section{Conclusion and Future Work}
In this paper, we introduced a hybrid framework that moves edge-cloud resource management from a reactive to proactive paradigm. We combined a CNN-LSTM predictive module with a deep reinforcement learning orchestrator, an architecture that overcomes the failures of the traditional approaches directly.

Our key innovation consisted of feeding the DRL agent a ``lookahead'' state space. It is this simple change that enables the agent to make anticipatory decisions and optimize for long-term performance, not just the current-state crisis. We are excited about a number of future directions. First and foremost, our work will focus on privacy: applying Federated Learning to train the predictive model on the edge nodes themselves—a method that avoids centralizing sensitive data. We will make our agent ``smarter'' about risk by developing an Uncertainty-Aware DRL agent. A Bayesian Neural Network would allow the agent to ``hedge'' against bad forecasts. Our ultimate objective is real-world deployment. We will explore Sim-to-Real Transfer, using domain randomization to train a policy that is robust enough to run on physical hardware.

\bibliography{all_references}

\end{document}